\documentclass[letterpaper]{article}
\usepackage{aaai25}
\usepackage{times}
\usepackage{helvet}
\usepackage{courier}
\usepackage[hyphens]{url}
\usepackage{graphicx}
\usepackage{natbib}
\usepackage{caption}
\usepackage{algorithm}
\usepackage[noend]{algpseudocode}
\usepackage{amsmath,amssymb,amsfonts}
\usepackage{booktabs}
\usepackage{multirow}
\usepackage[table]{xcolor}
\usepackage{subcaption}
\definecolor{improved}{gray}{0.88}

\newcommand{\gpix}{\mathbf{g}_{\mathrm{pix}}}
\newcommand{\gfreq}{\mathbf{g}_{\mathrm{freq}}}
\newcommand{\gfreqpara}{\mathbf{g}_{\mathrm{freq}}^{\parallel}}
\newcommand{\gfreqperp}{\mathbf{g}_{\mathrm{freq}}^{\perp}}
\newcommand{\hmgda}{\mathbf{h}_{\mathrm{mgda}}}
\newcommand{\htgrad}{\mathbf{h}_{t}}
\newcommand{\Rrect}{\mathcal{R}}
\newcommand{\wpix}{w_{\mathrm{pix}}}
\newcommand{\wfreq}{w_{\mathrm{freq}}}

\title{OGG-FR: Orthogonal Gradient Gaming and Frequency Rectification for Unmanned Aerial Vehicle Infrared Image Super-Resolution}

\author{
    Yongsong Huang\textsuperscript{\rm 1},
    Qingzhong Wang\textsuperscript{\rm 2},
    Xiaofeng Liu\textsuperscript{\rm 3},
    Tomo Miyazaki\textsuperscript{\rm 1}\\
    Yaohou Fan\textsuperscript{\rm 1},
    Shinichiro Omachi\textsuperscript{\rm 1}
}
\affiliations{
    \textsuperscript{\rm 1}Tohoku University,
    \textsuperscript{\rm 2}Amazon Web Services,
    \textsuperscript{\rm 3}Yale University\\
    (hys, tomo, fan.yaohou.t4, shinichiro.omachi.b5)@tohoku.ac.jp,
    qzwang@amazon.com, xiaofeng.liu@yale.edu
}

\begin{document}
\maketitle

\begin{abstract}
Unmanned aerial vehicle (UAV) infrared image super-resolution aims to recover weak thermal structures for deployment on resource-constrained platforms; lightweight models are therefore preferred, but multi-loss training can be unstable. A common strategy combines pixel-domain and frequency-domain objectives; however, low contrast, limited high-frequency content, and sensor-specific noise often make their gradients weakly aligned or conflicting. To address this optimization ambiguity, we propose Orthogonal Gradient Gaming and Frequency Rectification (OGG-FR), a plug-and-play optimization framework that decomposes the frequency gradient into a redundant parallel component and an orthogonal innovation component relative to the pixel gradient. In the conflict regime, OGG-FR computes a safe base gradient using the Multiple Gradient Descent Algorithm (MGDA) and adds a variance-rectified orthogonal innovation; in the compatible regime, it discards redundant parallel information and injects the orthogonal innovation according to a confidence score estimated from the high-frequency residual. Experimental results on the UAV thermal benchmark show broad gains under BI and BD degradations at $\times 4$ and $\times 8$ scales, while gradient analyses support the effectiveness of the proposed conflict-aware update rule.
\end{abstract}

\begin{figure}[t]
\centerline{\includegraphics[width=0.85\columnwidth]{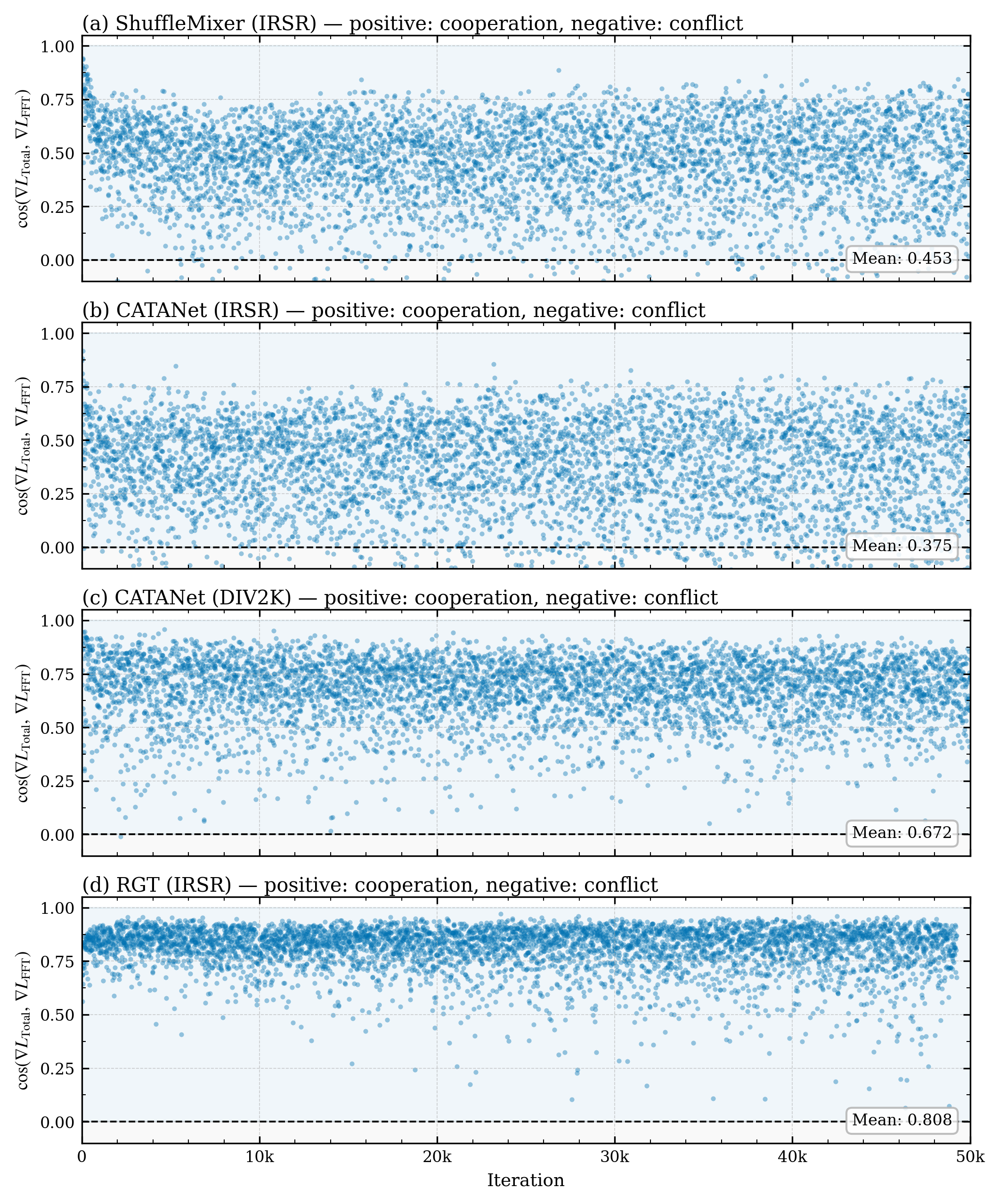}}
\caption{Gradient conflict is more pronounced in the compact infrared settings analyzed here but remains architecture-dependent. We visualize $\cos(\nabla L_{\mathrm{Total}}, \nabla L_{\mathrm{FFT}})$ during training, where positive and negative values indicate cooperation and conflict, respectively. The mean cosine similarities are 0.453 and 0.375 for the infrared settings in (a) and (b), compared with 0.672 for the visible-domain reference in (c) and 0.808 for the RGT infrared setting in (d).}
\label{fig:gradient_conflict}
\end{figure}

\section{Introduction}
\label{sec:introduction}

UAV infrared image super-resolution (IRSR) aims to reconstruct a high-resolution thermal image from a low-resolution infrared observation captured by an airborne platform. This task is important for nighttime surveillance, search and rescue, traffic monitoring, and remote sensing, where target visibility often depends on thermal contrast rather than color texture~\cite{chen2025fusion,barbato2026flyawarev2}. Recent UAV thermal benchmarks further show that low-light and adverse outdoor conditions make thermal reconstruction difficult because airborne thermal imaging is affected by limited sensor resolution, atmospheric interference, and weak texture cues~\cite{10938642,zhao2025guidance}. In practical UAV systems, the reconstruction model is usually deployed or adapted on resource-constrained edge devices, making lightweight super-resolution models more attractive than heavy restoration networks.

While lightweight architectures address the deployment constraint, stable training still depends on how complementary objectives are combined.
Modern lightweight SR training commonly combines complementary losses to balance pixel fidelity and perceptual detail. A pixel-domain loss such as $L_1$ provides stable supervision for global thermal structure, while an FFT loss encourages frequency consistency and high-frequency recovery. This combination is appealing for IRSR because infrared images contain large smooth thermal regions together with weak but task-critical boundaries. Huang et al.~\cite{11181143} further note that, compared with visible images, infrared images are characterized by lower contrast, less high-frequency detail, sensor-specific noise, and greater overlap between high- and low-frequency information. These properties make the interpretation of frequency-domain supervision ambiguous: a high-frequency response may correspond to a missing object boundary in one iteration, but to noise-like thermal fluctuation in another.

This ambiguity must therefore be addressed at the optimization level because $L_1$ and FFT losses are typically minimized using a fixed weighted sum. Such a sum is appropriate only when the objectives are broadly cooperative; when objectives compete, multi-objective optimization theory suggests that the update direction should explicitly model their trade-off~\cite{sener2018multi,wei2024mmpareto}. Our measurements show that this concern is not merely theoretical for UAV IRSR. As shown in Fig.~\ref{fig:gradient_conflict}, under the plotted backbones and training settings, infrared training exhibits lower loss-gradient cosine similarity and more frequent negative alignment than the visible-domain reference.

To address the above loss-gradient ambiguity and conflict in lightweight infrared SR training, we propose Orthogonal Gradient Gaming and Frequency Rectification (OGG-FR), a plug-and-play optimization framework designed for this infrared multi-loss training problem. OGG-FR treats the weighted pixel gradient $\gpix=\nabla_{\theta}(\wpix\mathcal{L}_{L_1})$ as the stable reconstruction direction and the weighted frequency gradient $\gfreq=\nabla_{\theta}(\wfreq\mathcal{L}_{\mathrm{FFT}})$ as a mixed signal containing both redundant and innovative components. It explicitly decomposes $\gfreq$ into the parallel component $\gfreqpara$, which is already covered by $\gpix$, and the orthogonal component $\gfreqperp$, which represents frequency-domain innovation unavailable to the pixel objective. This decomposition allows OGG-FR to preserve useful detail-seeking information without blindly trusting all frequency gradients.

OGG-FR is related to, but distinct from, generic gradient manipulation and loss-balancing methods. The Multiple Gradient Descent Algorithm (MGDA) computes a Pareto-stationary convex combination of task gradients~\cite{sener2018multi}, PCGrad removes pairwise conflicting components~\cite{yu2020gradient}, and adaptive balancing methods such as GradNorm~\cite{chen2018gradnorm}, CAGrad~\cite{liu2021conflict}, and FAMO~\cite{liu2023famo} regulate training through task-level gradient geometry or progress. These methods are general-purpose, whereas OGG-FR exploits the image-specific interpretation of the FFT gradient: the parallel component is treated as redundant or conflicting, and the orthogonal component is retained only after conflict-aware or residual-aware rectification. This design is especially important for infrared SR, where high-frequency responses can represent either weak thermal boundaries or sensor-induced fluctuations.

The proposed update rule is governed by the gradient gaming cosine $\rho_t$. We use the natural zero boundary throughout: when $\rho_t<0$, OGG-FR enters a destructive conflict regime, computes a Pareto-safe base gradient by MGDA, and adds a variance-rectified $\gfreqperp$ to retain robust frequency innovation while suppressing unstable high-variance components. When $\rho_t\geq0$, OGG-FR enters a compatible regime: it discards the redundant $\gfreqpara$ and uses a high-frequency residual confidence $s_t$ to decide how strongly $\gfreqperp$ should guide the update. In this way, the same framework handles both conflict control and detail recovery.

Our contributions are summarized as follows:
\begin{itemize}
    \item We identify gradient conflict between pixel-domain and frequency-domain objectives as a key optimization bottleneck for lightweight UAV infrared super-resolution, and show that the analyzed infrared settings exhibit stronger interference than the plotted visible-domain reference.
    \item We propose OGG-FR, a plug-and-play optimization framework that decomposes $\gfreq$ into $\gfreqpara$ and $\gfreqperp$, separating redundant, destructive, and innovative frequency signals during training.
    \item We design a two-regime update rule: MGDA safe control with variance rectification for $\rho_t<0$, and residual-aware orthogonal calibration with $s_t$ for $\rho_t\geq0$.
    \item Experiments on the Low-light UAV thermal SR benchmark demonstrate broad improvements across representative lightweight and transformer-based SR models under BI/BD degradations and $\times4/\times8$ scales.
\end{itemize}

\section{Method}
\label{sec:method}

\begin{figure*}[t]
    \centering
    \includegraphics[width=\linewidth]{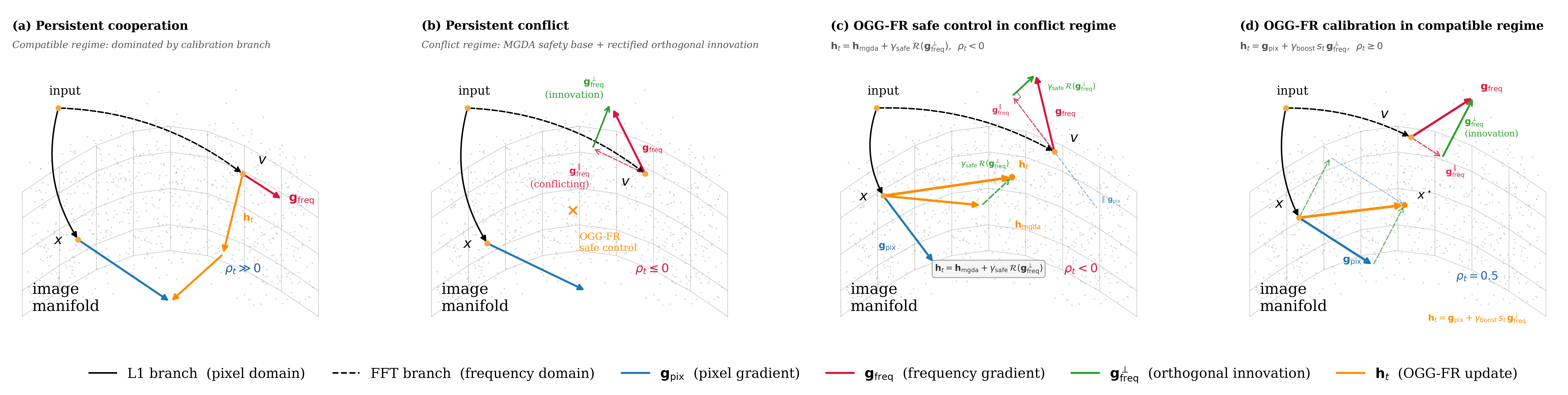}
    \caption{\textbf{Overview of the proposed Orthogonal Gradient Gaming and Frequency Rectification (OGG-FR) framework.}
    \textbf{(a)} Persistent cooperation: well-aligned $\gpix$ and $\gfreq$ keep optimization in the compatible regime.
    \textbf{(b)} Persistent conflict: destructive disagreement is exposed by decomposing $\gfreq$ into $\gfreqpara$ and $\gfreqperp$.
    \textbf{(c)} Safe control: for $\rho_t<0$ (Case A), OGG-FR forms $\htgrad=\hmgda+\gamma_{\mathrm{safe}}\Rrect(\gfreqperp)$.
    \textbf{(d)} Calibration: for $\rho_t\geq0$ (Case B), OGG-FR discards $\gfreqpara$ and uses $\htgrad=\gpix+\gamma_{\mathrm{boost}}s_t\gfreqperp$.
    Here, $\gfreqpara=\text{Proj}_{\gpix}(\gfreq)$ and $\gfreqperp=\gfreq-\gfreqpara$.}
    \label{fig:oggfr_overview}
\end{figure*}

The OGG-FR training update is organized around the geometry between the pixel-domain gradient and the frequency-domain gradient, as illustrated in Fig.~\ref{fig:oggfr_overview}. The key idea is to avoid treating the FFT gradient as a single indivisible signal: after measuring the gaming cosine $\rho_t$, OGG-FR separates frequency information that is already explained by $\gpix$ from orthogonal frequency innovation that may still improve detail recovery. The update is then selected according to the regime, using safe control when the two objectives conflict and residual-aware calibration when they are compatible. Algorithm~\ref{alg:oggfr} gives the complete training step.

\subsection{Problem Setup}
Let $\mathbf{x}\in\mathcal{X}\subset\mathbb{R}^{h\times w}$ denote a low-resolution infrared image and $\mathbf{y}\in\mathcal{Y}\subset\mathbb{R}^{H\times W}$ denote its high-resolution target, where $H=sh$ and $W=sw$ for scale factor $s$. A super-resolution model $f_{\theta}$ predicts $\hat{\mathbf{y}}=f_{\theta}(\mathbf{x})$. We train the model using a pixel loss and a frequency loss:
\begin{equation}
\mathcal{L}_{L_1}=\|\hat{\mathbf{y}}-\mathbf{y}\|_1,
\quad
\mathcal{L}_{\mathrm{FFT}}=
\left\|\left|\mathcal{F}(\hat{\mathbf{y}})\right|-
\left|\mathcal{F}(\mathbf{y})\right|\right\|_1,
\end{equation}
where $\mathcal{F}(\cdot)$ denotes the 2D FFT. At iteration $t$, the corresponding gradients are
\begin{equation}
\gpix=\nabla_{\theta}\left(\wpix\mathcal{L}_{L_1}\right),
\qquad
\gfreq=\nabla_{\theta}\left(\wfreq\mathcal{L}_{\mathrm{FFT}}\right),
\end{equation}
where $\wpix$ and $\wfreq$ are the loss weights used in the baseline training objective. The standard weighted-sum update implicitly assumes that $\gpix$ and $\gfreq$ are compatible. OGG-FR instead checks their geometric relation before composing the final update gradient $\htgrad$.

\subsection{Orthogonal Gradient Gaming}
OGG-FR uses the gradient gaming cosine $\rho_t$ to measure whether the pixel and frequency objectives cooperate:
\begin{equation}
\rho_t=
\frac{\gpix^{\top}\gfreq}
{\|\gpix\|\|\gfreq\|}.
\end{equation}
We fix the regime boundary at zero throughout the paper. Thus, negative cosine similarity indicates destructive disagreement between pixel and frequency objectives, while non-negative cosine similarity indicates compatible descent.

The core operation is to decompose $\gfreq$ with respect to $\gpix$:
\begin{equation}
\gfreqpara=
\frac{\gpix^{\top}\gfreq}{\|\gpix\|^2}\gpix,
\qquad
\gfreqperp=\gfreq-\gfreqpara.
\end{equation}
The parallel component $\gfreqpara$ represents the part of the FFT update that lies on the pixel-gradient axis; it is redundant when it agrees with $\gpix$ and unsafe when the two objectives oppose each other. The orthogonal component $\gfreqperp$ contains information not explained by the pixel-domain objective, so OGG-FR treats it as the main frequency-domain innovation signal.

\begin{algorithm}[t]
\caption{OGG-FR Training Update}
\label{alg:oggfr}
\begin{algorithmic}[1]
\Require Model parameters $\theta_t$, batch data, learning rate $\mu$, coefficients $\gamma_{\mathrm{boost}},\gamma_{\mathrm{safe}},\beta,\kappa,\eta_{\mathrm{cut}}$
\Ensure Updated parameters $\theta_{t+1}$
\State Compute $\gpix\leftarrow\nabla_{\theta}(\wpix\mathcal{L}_{L_1})$ and $\gfreq\leftarrow\nabla_{\theta}(\wfreq\mathcal{L}_{\mathrm{FFT}})$ independently
\State Compute $\rho_t\leftarrow\gpix^{\top}\gfreq/(\|\gpix\|\|\gfreq\|)$
\State Decompose $\gfreqpara\leftarrow(\gpix^{\top}\gfreq/\|\gpix\|^2)\gpix$ and $\gfreqperp\leftarrow\gfreq-\gfreqpara$
\State Compute $s_t\leftarrow\sigma(\kappa(\mathcal{S}(\mathbf{r}_{H})-1))$
\If{$\rho_t<0$}
    \State Compute $\alpha^{*}$ and $\hmgda$
    \State $\htgrad\leftarrow\hmgda+\gamma_{\mathrm{safe}}\Rrect(\gfreqperp)$
\Else
    \State $\htgrad\leftarrow\gpix+\gamma_{\mathrm{boost}}s_t\gfreqperp$
\EndIf
\State $\theta_{t+1}\leftarrow\theta_t-\mu\htgrad$
\end{algorithmic}
\end{algorithm}

\subsection{Case A: Safe Control in the Conflict Regime}
When $\rho_t<0$, directly summing $\gpix$ and $\gfreq$ can produce an update that damages pixel fidelity or oscillates between objectives. OGG-FR first computes an MGDA safe base gradient:
\begin{equation}
\hmgda=\alpha^{*}\gpix+(1-\alpha^{*})\gfreq,
\end{equation}
where the closed-form coefficient is
\begin{equation}
\alpha^{*}=
\operatorname{clip}
\left(
\frac{(\gfreq-\gpix)^{\top}\gfreq}
{\|\gfreq-\gpix\|^{2}},
0,1
\right).
\end{equation}
Here, MGDA chooses $\alpha^{*}$ so that $\hmgda$ is the minimum-norm point in the convex hull of $\{\gpix,\gfreq\}$, i.e.,
$\alpha^{*}=\arg\min_{\alpha\in[0,1]}\|\alpha\gpix+(1-\alpha)\gfreq\|_2^2$.
This construction yields a Pareto-safe compromise direction under gradient conflict: instead of following either $\gpix$ or $\gfreq$ alone, $\hmgda$ balances the two objectives while reducing the risk of a destructive update. However, a pure compromise may discard useful high-frequency innovation. We therefore add a rectified orthogonal component:
\begin{equation}
\Rrect(\gfreqperp)=
\frac{\gfreqperp}
{1+\beta\operatorname{Var}(\gfreqperp)},
\end{equation}
where $\operatorname{Var}(\cdot)$ denotes the variance of the flattened gradient vector and $\beta$ controls the strength of variance suppression. The final conflict-regime update is
\begin{equation}
\htgrad=\hmgda+\gamma_{\mathrm{safe}}\Rrect(\gfreqperp).
\end{equation}
High variance in $\gfreqperp$ indicates unstable frequency responses, so the rectifier suppresses the orthogonal component; low variance indicates more coherent orthogonal innovation, so the component is retained.

\subsection{Case B: Residual-Aware Calibration in the Compatible Regime}
When $\rho_t\geq0$, the two objectives are directionally compatible, but the parallel part of $\gfreq$ is still redundant because $\gpix$ already provides the corresponding descent direction. OGG-FR therefore discards $\gfreqpara$ and adds only $\gfreqperp$:
\begin{equation}
\htgrad=\gpix+\gamma_{\mathrm{boost}}s_t\gfreqperp.
\end{equation}
The scalar $s_t\in(0,1)$ estimates how much residual high-frequency activity should be trusted when injecting the orthogonal frequency innovation.

To compute $s_t$, we first define the residual $\mathbf{r}=\hat{\mathbf{y}}-\mathbf{y}$ and extract its high-frequency component:
\begin{equation}
\mathbf{r}_{H}=
\mathcal{F}^{-1}
\left(
\mathcal{F}(\mathbf{r})\odot\mathbf{M}_{H}
\right),
\end{equation}
where $\mathbf{M}_{H}$ is a high-frequency mask whose cutoff condition is $f_r>\eta_{\mathrm{cut}}f_{\mathrm{Nyquist}}$. We then measure the spatial structure of the high-frequency residual by
\begin{equation}
\mathcal{S}(\mathbf{r}_{H})=
\frac{\mathrm{TV}(\mathbf{r}_{H})}
{\|\mathbf{r}_{H}\|_1+\varepsilon},
\end{equation}
with
\begin{equation}
\begin{aligned}
\mathrm{TV}(\mathbf{r}_{H})=
\sum_{i,j}\big(
&\left|[\mathbf{r}_{H}]_{i,j+1}-[\mathbf{r}_{H}]_{i,j}\right| \\
&+\left|[\mathbf{r}_{H}]_{i+1,j}-[\mathbf{r}_{H}]_{i,j}\right|
\big).
\end{aligned}
\end{equation}
We use $\mathcal{S}(\mathbf{r}_{H})=1$ as a fixed normalization point rather than a universal edge/noise boundary. Larger values increase $s_t$ and allow stronger orthogonal frequency correction, while smaller values keep the update closer to $\gpix$. We use this fixed normalization point in all tested UAV settings; evaluating dataset-adaptive thresholds remains an important direction for future work.

\begin{table*}[t]
\centering
\renewcommand\arraystretch{1.1}
\caption{
Average IRSR performance in terms of PSNR$\uparrow$, MSE$\downarrow$, SSIM$\uparrow$, and NIQE$\downarrow$ on UAV-BD and UAV-BI at scale factors $\times 4$ and $\times 8$ after approximately 500 epochs. Shaded cells indicate improvements over the baseline.
}
\resizebox{\textwidth}{!}{\begin{tabular}{@{}c|c|c|c|cccc|cccc@{}}
\toprule
\multirow{2}{*}{Dataset}
& \multirow{2}{*}{Scale}
& \multirow{2}{*}{Methods}
& \multicolumn{1}{c|}{\multirow{2}{*}{\# Params.~(K)}}
& \multicolumn{4}{c|}{\text{w/o OGG-FR}}
& \multicolumn{4}{c}{\text{OGG-FR}} \\
\cmidrule(l){5-8}\cmidrule(l){9-12}
& & &
& PSNR$\uparrow$ & MSE$\downarrow$ & SSIM$\uparrow$ & NIQE$\downarrow$
& PSNR$\uparrow$ & MSE$\downarrow$ & SSIM$\uparrow$ & NIQE$\downarrow$ \\
\midrule
\multirow{16}{*}{UAV-BD}
  & \multirow{9}{*}{$\times 4$}
  & ShuffleMixer~\cite{sun2022shufflemixer} & 108
  & 30.4104 & 76.3041 & 0.8579 & 7.3936
  & \cellcolor{improved}31.2194 & \cellcolor{improved}70.3025 & \cellcolor{improved}0.8786 & \cellcolor{improved}5.5961 \\
  & & ShuffleMixer (base)~\cite{sun2022shufflemixer} & 121
  & 33.8813 & 43.1739 & 0.9187 & 5.7070
  & \cellcolor{improved}34.5711 & \cellcolor{improved}36.5033 & \cellcolor{improved}0.9289 & 5.7794 \\
  & & CRAFT~\cite{li2023craft} & 900
  & 33.3686 & 44.8310 & 0.9140 & 5.9005
  & \cellcolor{improved}33.6179 & \cellcolor{improved}43.9027 & \cellcolor{improved}0.9164 & \cellcolor{improved}5.8298 \\
  & & CATANet~\cite{liu2025catanet} & 477
  & 32.8713 & 52.6394 & 0.9054 & 5.4843
  & \cellcolor{improved}33.4070 & \cellcolor{improved}47.3185 & \cellcolor{improved}0.9132 & 5.8639 \\
  & & SMFANet~\cite{luo2024smfanet} & 197
  & 30.4008 & 89.5197 & 0.8599 & 6.1684
  & \cellcolor{improved}30.4494 & \cellcolor{improved}89.0369 & \cellcolor{improved}0.8617 & \cellcolor{improved}5.8442 \\
  & & SMFANet$^{+}$~\cite{luo2024smfanet} & 496
  & 30.3668 & 89.0442 & 0.8609 & 6.1848
  & \cellcolor{improved}30.4892 & \cellcolor{improved}87.7844 & \cellcolor{improved}0.8633 & \cellcolor{improved}5.8833 \\
  & & HiT~\cite{chen2023hit} & 792
  & 35.2325 & 31.6830 & 0.9373 & 5.5530
  & \cellcolor{improved}35.4725 & \cellcolor{improved}29.1665 & \cellcolor{improved}0.9388 & 5.7654 \\
  & & SwinIR~\cite{liang2021swinir} & 11{,}752
  & 34.8334 & 32.9221 & 0.9339 & 5.7792
  & \cellcolor{improved}35.3735 & \cellcolor{improved}29.1706 & \cellcolor{improved}0.9390 & 5.8153 \\
  & & RGT~\cite{chen2024recursive} & 10{,}051
  & 34.2171 & 36.6447 & 0.9282 & 5.4614
  & 34.0957 & 38.7636 & 0.9248 & 5.9451 \\
  & & HAT~\cite{Chen_2023_CVPR} & 20{,}624
  & 33.2487 & 46.0882 & 0.9147 & 5.5795
  & \cellcolor{improved}33.4802 & \cellcolor{improved}44.2329 & \cellcolor{improved}0.9152 & 5.8217 \\
\cmidrule(l){2-12}
  & \multirow{7}{*}{$\times 8$}
  & ShuffleMixer~\cite{sun2022shufflemixer} & 108
  & 27.1116 & 176.2158 & 0.7667 & 8.4598
  & \cellcolor{improved}27.3156 & \cellcolor{improved}169.0754 & \cellcolor{improved}0.7733 & \cellcolor{improved}8.0811 \\
  & & ShuffleMixer (base)~\cite{sun2022shufflemixer} & 121
  & 27.0887 & 178.0644 & 0.7638 & 6.4348
  & \cellcolor{improved}27.4328 & \cellcolor{improved}169.6041 & \cellcolor{improved}0.7757 & \cellcolor{improved}6.1854 \\
  & & CATANet~\cite{liu2025catanet} & 477
  & 27.9993 & 149.5698 & 0.7993 & 6.3305
  & \cellcolor{improved}28.4881 & \cellcolor{improved}136.0331 & \cellcolor{improved}0.8069 & 6.4007 \\
  & & SMFANet~\cite{luo2024smfanet} & 197
  & 26.8039 & 192.5686 & 0.7533 & 7.5856
  & \cellcolor{improved}26.8811 & \cellcolor{improved}189.7031 & \cellcolor{improved}0.7578 & \cellcolor{improved}6.3541 \\
  & & SMFANet$^{+}$~\cite{luo2024smfanet} & 496
  & 26.6788 & 187.7343 & 0.7600 & 6.6701
  & \cellcolor{improved}27.0244 & \cellcolor{improved}184.4603 & \cellcolor{improved}0.7635 & \cellcolor{improved}6.2231 \\
  & & CRAFT~\cite{li2023craft} & 900
  & 27.2763 & 174.3905 & 0.7724 & 6.6143
  & \cellcolor{improved}27.3799 & \cellcolor{improved}171.7478 & \cellcolor{improved}0.7732 & \cellcolor{improved}6.3427 \\
  & & HiT~\cite{chen2023hit} & 792
  & 28.0913 & 145.4324 & 0.7909 & 6.2176
  & \cellcolor{improved}28.5454 & \cellcolor{improved}132.1671 & \cellcolor{improved}0.8066 & 6.3945 \\
\midrule
\multirow{17}{*}{UAV-BI}
  & \multirow{10}{*}{$\times 4$}
  & ShuffleMixer~\cite{sun2022shufflemixer} & 108
  & 33.7076 & 45.6251 & 0.9112 & 5.2927
  & \cellcolor{improved}34.0009 & \cellcolor{improved}43.0788 & \cellcolor{improved}0.9175 & 5.8163 \\
  & & ShuffleMixer (base)~\cite{sun2022shufflemixer} & 121
  & 34.8496 & 35.3212 & 0.9302 & 5.8230
  & 34.7982 & 35.8350 & 0.9292 & \cellcolor{improved}5.7946 \\
  & & CATANet~\cite{liu2025catanet} & 477
  & 34.4801 & 38.3136 & 0.9261 & 5.6563
  & \cellcolor{improved}34.5207 & \cellcolor{improved}38.0689 & 0.9255 & 5.8342 \\
  & & SMFANet~\cite{luo2024smfanet} & 197
  & 32.3689 & 60.7782 & 0.8892 & 5.5994
  & \cellcolor{improved}32.4129 & \cellcolor{improved}59.6147 & \cellcolor{improved}0.8914 & 5.7438 \\
  & & SMFANet$^{+}$~\cite{luo2024smfanet} & 496
  & 32.6384 & 57.4679 & 0.8941 & 5.5590
  & \cellcolor{improved}32.8036 & \cellcolor{improved}55.4282 & \cellcolor{improved}0.8980 & 5.9040 \\
  & & CRAFT~\cite{li2023craft} & 900
  & 34.4442 & 37.8287 & 0.9251 & 5.3806
  & \cellcolor{improved}34.5370 & \cellcolor{improved}36.8635 & \cellcolor{improved}0.9271 & 5.9019 \\
  & & HiT~\cite{chen2023hit} & 792
  & 33.8040 & 43.3181 & 0.9151 & 5.2674
  & 33.7178 & 43.5052 & \cellcolor{improved}0.9159 & 5.6895 \\
  & & HAT~\cite{Chen_2023_CVPR} & 20{,}624
  & 33.3977 & 46.0394 & 0.9106 & 5.6089
  & \cellcolor{improved}34.3125 & \cellcolor{improved}37.1775 & \cellcolor{improved}0.9261 & 5.9253 \\
  & & RGT~\cite{chen2024recursive} & 10{,}051
  & 34.9598 & 32.5352 & 0.9330 & 5.5579
  & \cellcolor{improved}35.0247 & \cellcolor{improved}31.4878 & \cellcolor{improved}0.9340 & 5.9439 \\
  & & SwinIR~\cite{liang2021swinir} & 11{,}752
  & 35.1393 & 31.1943 & 0.9365 & 5.7304
  & 34.8297 & 31.7060 & 0.9353 & 5.8948 \\
\cmidrule(l){2-12}
  & \multirow{7}{*}{$\times 8$}
  & ShuffleMixer~\cite{sun2022shufflemixer} & 108
  & 28.2310 & 137.9232 & 0.7834 & 5.5137
  & \cellcolor{improved}28.8671 & \cellcolor{improved}125.4894 & \cellcolor{improved}0.7969 & 5.6093 \\
  & & ShuffleMixer (base)~\cite{sun2022shufflemixer} & 121
  & 28.4231 & 134.8031 & 0.7871 & 6.3474
  & \cellcolor{improved}28.4965 & \cellcolor{improved}131.1873 & \cellcolor{improved}0.7899 & \cellcolor{improved}5.4892 \\
  & & CATANet~\cite{liu2025catanet} & 477
  & 29.3685 & 115.7594 & 0.8168 & 6.2420
  & \cellcolor{improved}29.5680 & \cellcolor{improved}110.8248 & \cellcolor{improved}0.8199 & \cellcolor{improved}6.1149 \\
  & & SMFANet~\cite{luo2024smfanet} & 197
  & 28.1988 & 142.9715 & 0.7771 & 5.9971
  & \cellcolor{improved}28.5738 & \cellcolor{improved}133.7659 & \cellcolor{improved}0.7898 & 6.2159 \\
  & & SMFANet$^{+}$~\cite{luo2024smfanet} & 496
  & 28.0576 & 145.4552 & 0.7759 & 6.0762
  & 28.0344 & 147.3876 & \cellcolor{improved}0.7764 & 6.2181 \\
  & & CRAFT~\cite{li2023craft} & 900
  & 29.4707 & 111.2567 & 0.8146 & 6.4938
  & 29.3067 & 113.4951 & 0.8128 & \cellcolor{improved}6.3606 \\
  & & HiT~\cite{chen2023hit} & 792
  & 29.9354 & 100.7407 & 0.8266 & 6.1823
  & \cellcolor{improved}29.9530 & \cellcolor{improved}100.7364 & \cellcolor{improved}0.8295 & \cellcolor{improved}6.1658 \\
\bottomrule
\end{tabular}}
\label{tab:irsr_results_200epoch}
\end{table*}

\section{Experiments}
\label{sec:experiments}

We evaluate OGG-FR from both reconstruction and optimization perspectives. The experiments first describe the UAV infrared benchmark and training protocol, then compare OGG-FR with standard weighted-sum training across representative SR backbones. We further analyze gradient conflict dynamics, ablate the rectification hyperparameters, and provide visual comparisons to connect the optimization behavior with reconstructed thermal details.

\begin{table*}[t]
\centering
\renewcommand\arraystretch{1.1}
\caption{
Visible-domain reference results on DIV2K at scale factor $\times 4$.
\colorbox{improved}{Shaded cells} indicate improvement over the baseline.
}
\resizebox{\textwidth}{!}{\begin{tabular}{@{}c|c|c|c|cccc|cccc@{}}
\toprule
\multirow{2}{*}{Dataset}
& \multirow{2}{*}{Scale}
& \multirow{2}{*}{Methods}
& \multicolumn{1}{c|}{\multirow{2}{*}{\# Params.~(K)}}
& \multicolumn{4}{c|}{\text{w/o OGG-FR}}
& \multicolumn{4}{c}{\text{OGG-FR}} \\
\cmidrule(l){5-8}\cmidrule(l){9-12}
& & &
& PSNR$\uparrow$ & MSE$\downarrow$ & SSIM$\uparrow$ & NIQE$\downarrow$
& PSNR$\uparrow$ & MSE$\downarrow$ & SSIM$\uparrow$ & NIQE$\downarrow$ \\
\midrule
\multirow{5}{*}{DIV2K}
  & \multirow{5}{*}{$\times 4$}
  & ShuffleMixer~\cite{sun2022shufflemixer} & 108
  & 29.0337 & 125.2473 & 0.7969 & 5.8362
  & \cellcolor{improved}29.1432 & \cellcolor{improved}122.8583 & \cellcolor{improved}0.8007 & \cellcolor{improved}5.8334 \\
  & & ShuffleMixer (base)~\cite{sun2022shufflemixer} & 121
  & 29.8969 & 107.8078 & 0.8260 & 6.1071
  & \cellcolor{improved}30.0070 & \cellcolor{improved}105.7706 & \cellcolor{improved}0.8267 & \cellcolor{improved}5.9743 \\
  & & SMFANet~\cite{luo2024smfanet} & 197
  & 28.5484 & 139.4078 & 0.7916 & 6.2981
  & \cellcolor{improved}28.5979 & \cellcolor{improved}137.7289 & \cellcolor{improved}0.7936 & 6.4677 \\
  & & HiT~\cite{chen2023hit} & 792
  & 29.0356 & 127.0732 & 0.8067 & 5.4705
  & \cellcolor{improved}29.4685 & \cellcolor{improved}115.4948 & \cellcolor{improved}0.8184 & 5.9140 \\
  & & RGT~\cite{chen2024recursive} & 10{,}051
  & 29.5132 & 116.1365 & 0.8165 & 6.2523
  & \cellcolor{improved}29.6875 & \cellcolor{improved}110.1185 & \cellcolor{improved}0.8236 & 6.2626 \\
\bottomrule
\end{tabular}}
\label{tab:div2k_reference}
\end{table*}

\subsection{Experimental Setup}
We evaluate OGG-FR on the Low-light split of the UAV thermal super-resolution benchmark introduced by Zhao et al.~\cite{zhao2025guidance}. The split contains separate training and testing folders with ground-truth thermal images, high-resolution RGB images, and low-resolution thermal inputs. Specifically, the Low-light split contains 925 training images and 237 testing images, each paired with a ground-truth thermal image and a high-resolution RGB image. The low-resolution thermal inputs include both bicubic interpolation (BI) and blur-downsampling (BD) degradations, corresponding to the UAV-BI and UAV-BD settings used in our tables; for each degradation, the dataset provides $\times 4$ and $\times 8$ low-resolution inputs, giving 925 training and 237 testing samples per degradation-scale setting. Following the experimental protocol, we reported UAV results at $\times 4$ and $\times 8$ scales and included DIV2K~\cite{agustsson2017ntire} at $\times 4$ as a visible-domain reference for gradient behavior. We restricted the DIV2K reference to $\times 4$ because the $\times 8$ setting was used primarily to stress-test the UAV infrared benchmark. For hierarchical backbones such as SwinIR and RGT, repeated downsampling can reduce already small infrared inputs below the models' minimum spatial-resolution requirements.

The evaluation metrics are PSNR, MSE, SSIM, and NIQE~\cite{niqe}. Higher PSNR and SSIM indicate better fidelity, while lower MSE and NIQE indicate lower distortion and better perceptual naturalness. All models are implemented in PyTorch and trained on NVIDIA A6000 GPUs. We use the Adam optimizer~\cite{kingma2014adam} for all backbones, with the learning rate and batch size following the corresponding training configuration: most transformer and hybrid models use a learning rate of $2\times10^{-4}$, and the batch size per GPU is set to 32 depending on the memory footprint of the backbone.

OGG-FR was evaluated as an optimizer-level plug-in rather than as a new architecture. We applied it to representative lightweight and transformer-based SR models, including ShuffleMixer, SMFANet, CATANet, CRAFT, HiT, SwinIR, RGT, and HAT, for which paired baseline and OGG-FR training records were available. Unless otherwise specified, the baseline weighted objective used $\wpix=1.0$ for $\mathcal{L}_{L_1}$ and $\wfreq=0.01$ for $\mathcal{L}_{\mathrm{FFT}}$; OGG-FR used the same loss weights when computing $\gpix$ and $\gfreq$. We set the fixed boundary to $\rho_t=0$ and used $\eta_{\mathrm{cut}}=0.5$, $\kappa=10$, $\gamma_{\mathrm{boost}}=1.0$, $\gamma_{\mathrm{safe}}=0.5$, and $\beta=1.0$. OGG-FR changes only the training update and adds no inference-time parameters or FLOPs; its extra cost comes from computing loss-specific gradients and lightweight FFT/TV statistics during training.

\subsection{Main Quantitative Results}
Table~\ref{tab:irsr_results_200epoch} reports average IRSR performance after approximately 500 training epochs. It compares each backbone trained with the standard weighted objective with the same backbone trained using OGG-FR, thereby isolating the effect of the optimization strategy from that of the architecture. OGG-FR improves many model-setting pairs, particularly in the challenging $\times 8$ UAV settings, where frequency recovery is more difficult and weighted-sum optimization is less stable. The largest PSNR gain is observed for HAT on UAV-BI at $\times 4$, where OGG-FR improves PSNR from 33.3977 dB to 34.3125 dB and SSIM from 0.9106 to 0.9261. Lightweight backbones also benefit: the 108K-parameter ShuffleMixer gains 0.8090 dB on UAV-BD at $\times 4$ and 0.6361 dB on UAV-BI at $\times 8$, showing that the proposed update can benefit capacity-limited models without adding inference cost.

Compared with the DIV2K $\times 4$ visible-domain reference reported separately in Table~\ref{tab:div2k_reference}, where the gains are consistently smaller, the UAV infrared settings show larger and more frequent improvements, which is consistent with the stronger loss-gradient ambiguity observed in Fig.~\ref{fig:gradient_conflict}. The gains are not limited to a single architecture family: improvements appear in convolutional lightweight models and transformer-style models, supporting the claim that OGG-FR addresses an optimization issue rather than an architecture-specific weakness.

\begin{figure*}[t]
    \centering
    \begin{subfigure}[b]{0.49\textwidth}
        \centering
        \includegraphics[width=\textwidth]{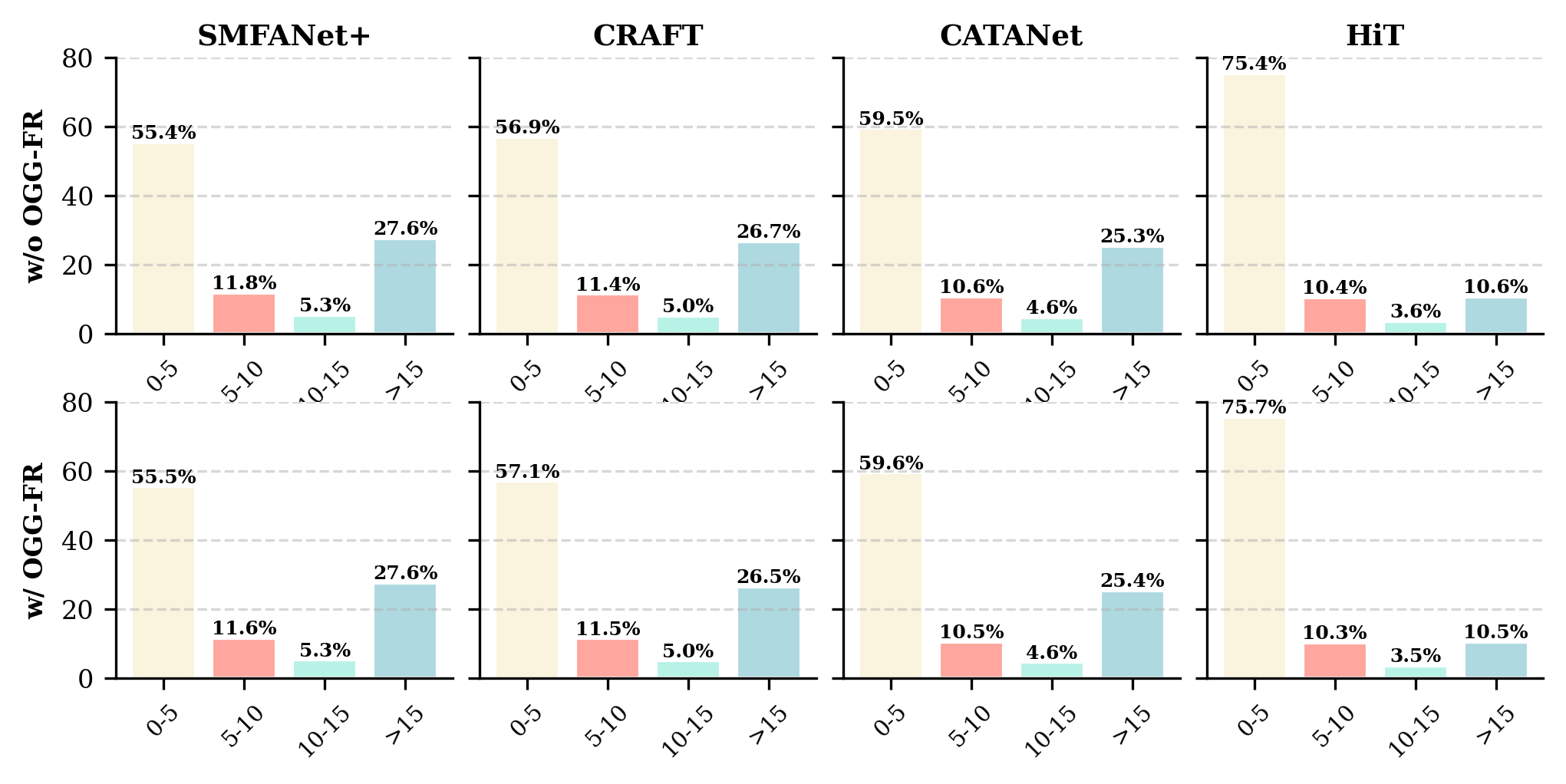}
        \caption{UAV-BI (Scale $\times 4$)}
        \label{fig:oggfr_bi_x4}
    \end{subfigure}
    \hfill
    \begin{subfigure}[b]{0.49\textwidth}
        \centering
        \includegraphics[width=\textwidth]{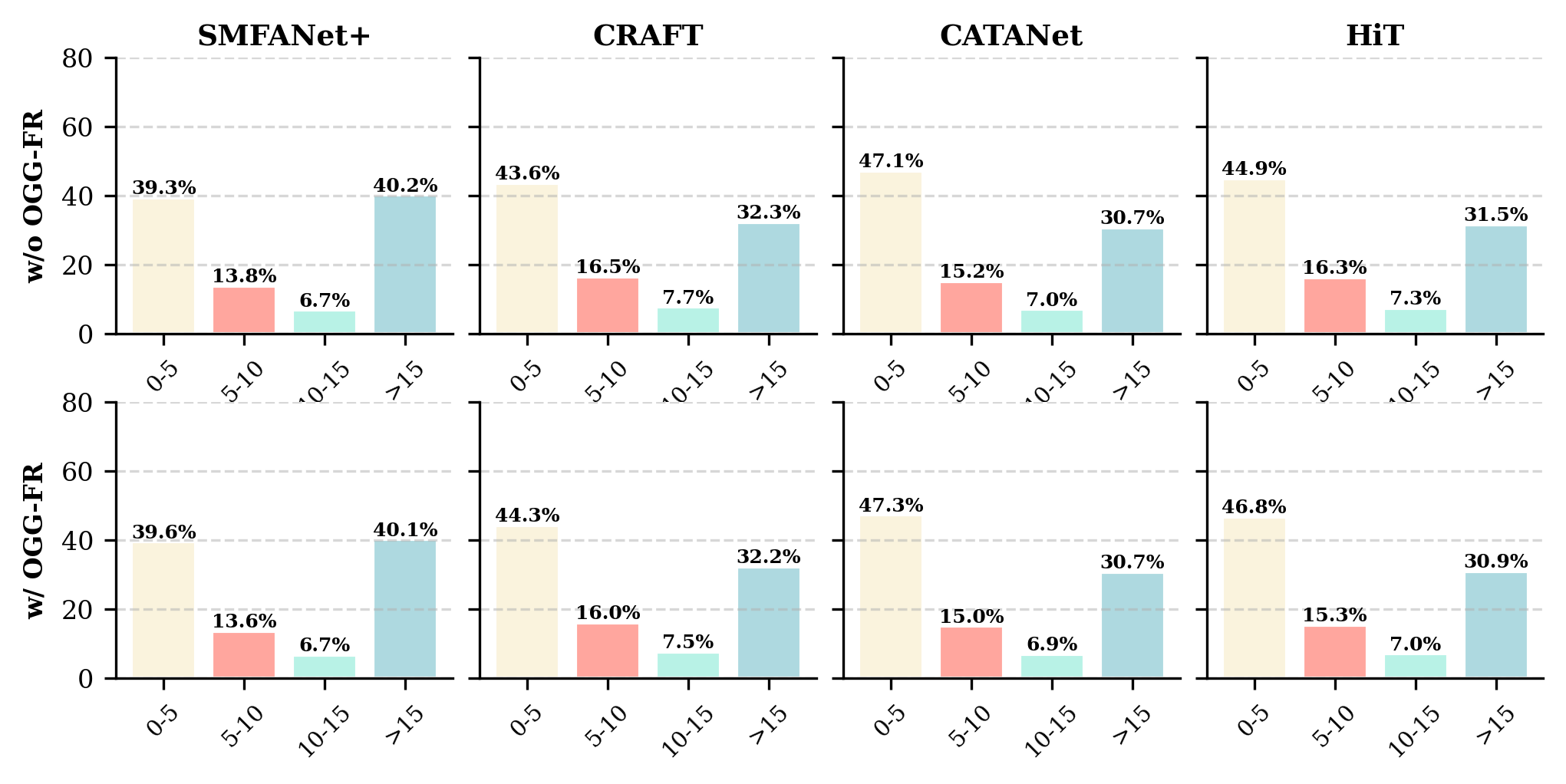}
        \caption{UAV-BI (Scale $\times 8$)}
        \label{fig:oggfr_bi_x8}
    \end{subfigure}
    \caption{Global error distribution analysis for various super-resolution models on the UAV-BI dataset at scaling factors of $\times 4$ and $\times 8$. The horizontal axis lists residual-error ranges from minimal to severe errors, and each paired bar compares the same backbone trained without and with OGG-FR.}
    \label{fig:oggfr_comparison}
\end{figure*}

\begin{figure}[t]
\centerline{\includegraphics[width=0.95\columnwidth]{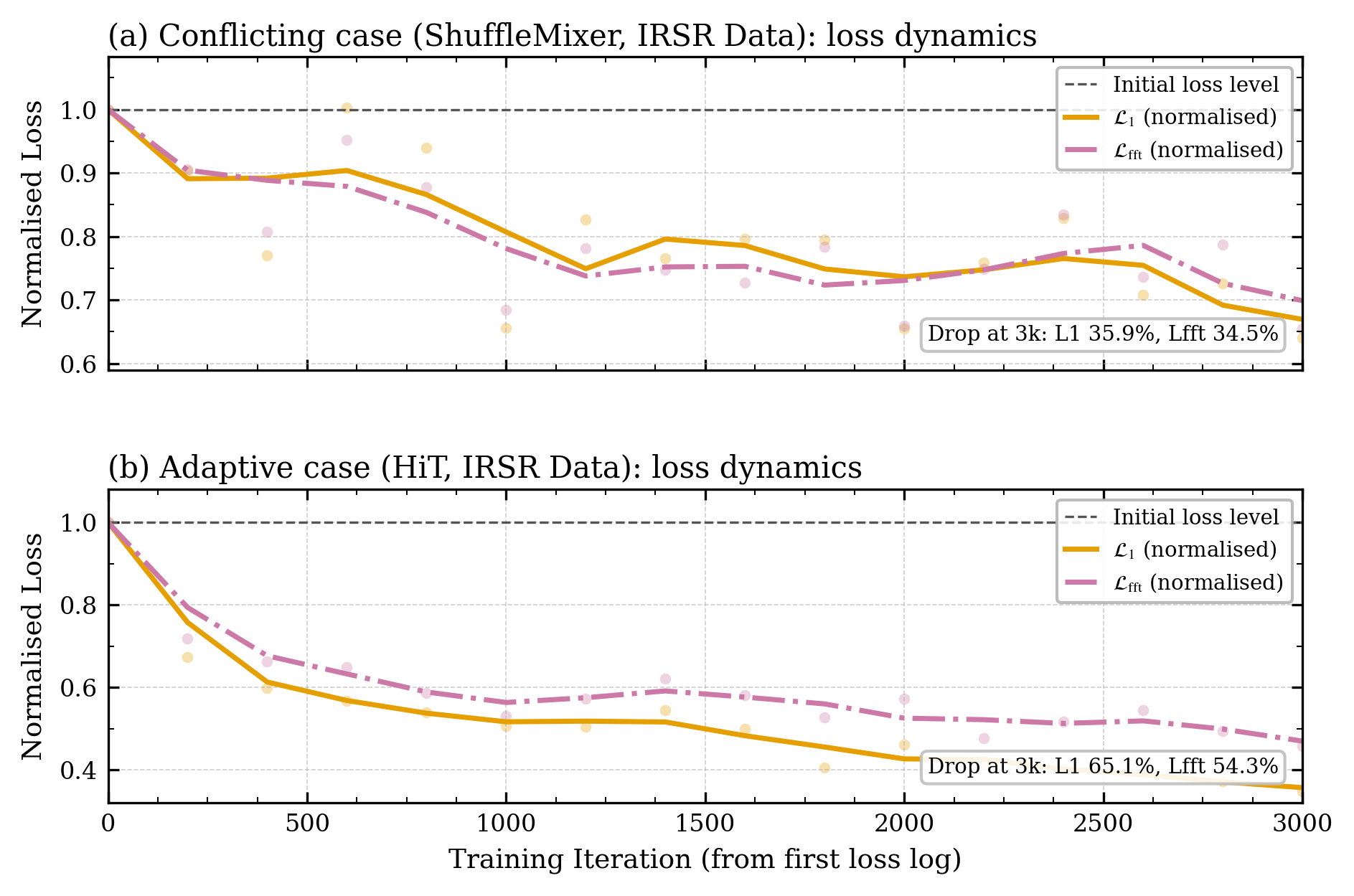}}
\caption{Training loss dynamics under OGG-FR optimization on IRSR data at scale $\times 4$. The curves report how the pixel-domain and frequency-domain objectives evolve during training, complementing Fig.~\ref{fig:ir_cosine} by showing that conflict-aware updates preserve stable descent while keeping frequency supervision active.}
\label{fig:loss_dynamics}
\end{figure}

We also report cases without improvement. In particular, the results for RGT on UAV-BD at $\times 4$, SwinIR on UAV-BI at $\times 4$, and SMFANet$^{+}$ on UAV-BI at $\times 8$ show that OGG-FR does not guarantee improvements for every architecture. Gains can be reduced or reversed when the baseline already provides stable frequency modeling or when the residual-confidence mechanism overweights weak orthogonal components.

Beyond average metrics, Fig.~\ref{fig:oggfr_comparison} examines the global error distribution before and after OGG-FR. The figure complements Table~\ref{tab:irsr_results_200epoch}: while the table reports aggregate fidelity and perceptual metrics, the error histogram shows whether the improvement comes from reducing severe reconstruction errors rather than only shifting well-reconstructed pixels. The reduction of high-error bins indicates that OGG-FR suppresses the tail of large reconstruction errors, which usually arise around weak thermal boundaries, small targets, and high-frequency residual regions. This supports its intended role as a stabilizer for difficult regions where direct weighted-sum training is more likely to inject noisy or conflicting frequency updates.

\subsection{Gradient Conflict Analysis}

Figure~\ref{fig:loss_dynamics} provides the corresponding optimization dynamics from the loss perspective. This figure is important for the evidence chain because a conflict-aware gradient rule should not only improve final metrics, but also avoid unstable training trajectories. The observed loss evolution supports the claim that the two-regime update keeps frequency supervision active while preventing destructive updates from dominating the pixel reconstruction path.

\begin{figure}[t]
\centerline{\includegraphics[width=\columnwidth]{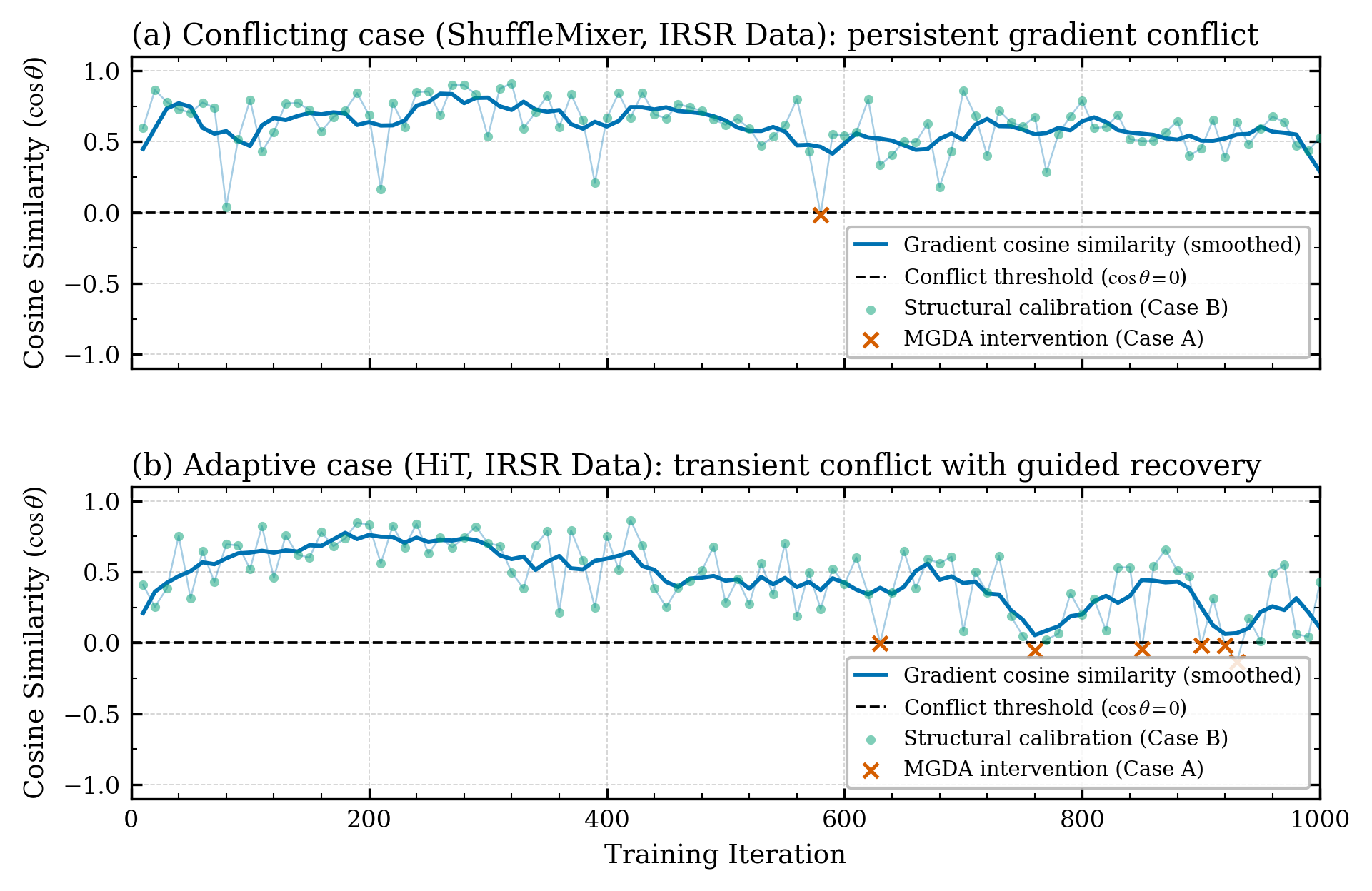}}
\caption{Gradient cosine similarity dynamics under OGG-FR optimization on IRSR data at scale $\times 4$. (a) ShuffleMixer exhibits persistent cooperation with rare MGDA intervention. (b) HiT shows transient conflicts in later iterations, triggering adaptive Case A control.}
\label{fig:ir_cosine}
\end{figure}

Figure~\ref{fig:ir_cosine} visualizes the training-time cosine similarity between pixel and frequency gradients on infrared data. The curves include many samples near zero or below zero, indicating that $L_1$ and FFT supervision are not consistently cooperative in IRSR. Together with the cross-domain comparison in Fig.~\ref{fig:gradient_conflict}, this validates the premise of OGG-FR: the frequency objective is useful, but its gradient should be interpreted before being merged into the update direction.

\begin{figure*}[t]
    \centering
    \includegraphics[width=1\textwidth]{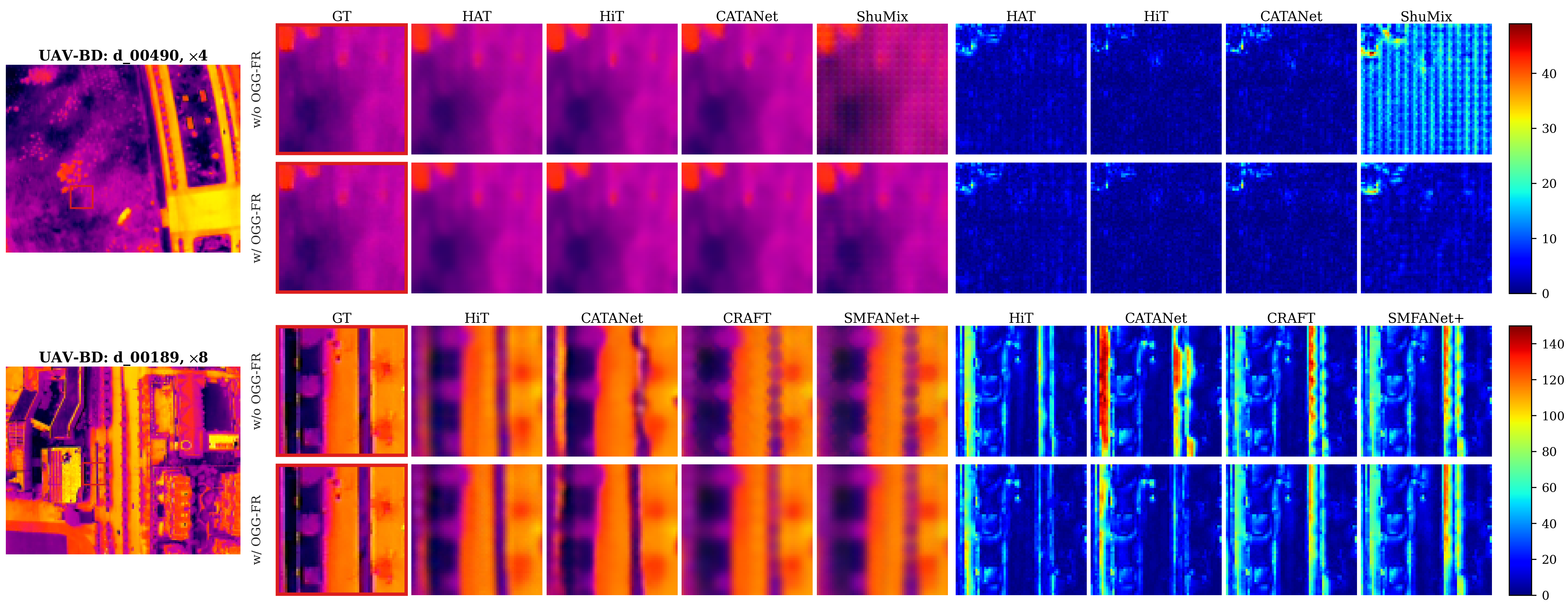}
    \caption{Visual comparison of $\times 4$ and $\times 8$ super-resolution results on UAV-BD images d\_00490 (top) and d\_00189 (bottom). With OGG-FR, the reconstructed thermal details are more faithful and better structured across different SR backbones, as indicated by the lower errors (darker blue) in the corresponding residual maps. Quantitatively, the mean crop residual decreases from 4.51 to 2.01 for d\_00490 at $\times 4$ and from 26.27 to 21.82 for d\_00189 at $\times 8$.}
    \label{fig:vis}
\end{figure*}

\subsection{Qualitative Results}
Figure~\ref{fig:vis} shows qualitative comparisons on UAV infrared scenes. This final visual evidence connects the optimization analysis back to the reconstruction target: OGG-FR tends to recover sharper thermal boundaries and more stable local structures, and the residual maps indicate smaller localized reconstruction errors. In the top example, d\_00490, OGG-FR preserves clearer object contours and reduces boundary residuals around the bright thermal target. In the bottom example, d\_00189, the reconstructed small structures are less blurred, and the residual maps show fewer concentrated error regions after applying OGG-FR. This behavior is consistent with the design of $\gfreqperp$: only the frequency component that is not already explained by $\gpix$ is injected, and its strength is controlled by either variance rectification or high-frequency residual structure.

\subsection{Ablation Studies}
We further ablate the two hyperparameters that directly control frequency rectification: the confidence slope $\kappa$ and the variance rectification coefficient $\beta$. Table~\ref{tab:ablation_kappa_beta} reports PSNR, MSE, and SSIM on three representative backbones under UAV-BI $\times 4$. The default $\kappa=10.0$ gives the best PSNR/SSIM for ShuffleMixer and HiT, while HAT remains nearly unchanged across $\kappa=5.0$ and $\kappa=20.0$, indicating that the confidence gate is not overly sensitive within this range. For $\beta$, the default $\beta=1.0$ provides a stable trade-off for ShuffleMixer and HiT, whereas overly strong variance rectification ($\beta=5.0$) clearly degrades HAT and reduces the gains on the other backbones. These results support using moderate confidence gating and moderate variance suppression: the former avoids underusing structured frequency residuals, while the latter avoids suppressing useful detail gradients. The fixed $\mathcal{S}(\mathbf{r}_H)=1$ inflection used by $s_t$ remains competitive in this setting, but the table also motivates future dataset-adaptive calibration. Additional sensitivity diagnostics of the update coefficients are provided in the supplementary material.

\begin{table}[t]
\centering
\renewcommand\arraystretch{1.1}
\caption{
Ablation study of $\kappa$ and $\beta$ on UAV-BI at a scale factor of $\times 4$.
}
\resizebox{0.9\columnwidth}{!}{\begin{tabular}{@{}c|c|c|ccc@{}}
\toprule
Ablation & Methods & Metric
& Low & Default & High \\
\midrule
\multirow{10}{*}{$\kappa$}
& \multicolumn{2}{c|}{}
& $\kappa=5.0$ & $\kappa=\mathbf{10.0}$ & $\kappa=20.0$ \\
\cmidrule(l){2-6}
& \multirow{3}{*}{ShuffleMixer}
  & PSNR$\uparrow$ & 34.4033 & \textbf{34.4669} & 34.4017 \\
& & MSE$\downarrow$ & 39.5385 & \textbf{38.7614} & 39.5676 \\
& & SSIM$\uparrow$ & 0.9231  & \textbf{0.9242}  & 0.9230  \\
\cmidrule(l){2-6}
& \multirow{3}{*}{HAT}
  & PSNR$\uparrow$ & 35.0330 & \textbf{35.0089} & 35.0298 \\
& & MSE$\downarrow$ & 31.4917 & \textbf{32.7994} & 31.7070 \\
& & SSIM$\uparrow$ & 0.9338  & \textbf{0.9328}  & 0.9333  \\
\cmidrule(l){2-6}
& \multirow{3}{*}{HiT}
  & PSNR$\uparrow$ & 35.5682 & \textbf{35.7043} & 35.5988 \\
& & MSE$\downarrow$ & 29.2823 & \textbf{28.2897} & 28.9719 \\
& & SSIM$\uparrow$ & 0.9389  & \textbf{0.9403}  & 0.9392  \\
\midrule
\multirow{10}{*}{$\beta$}
& \multicolumn{2}{c|}{}
& $\beta=0.5$ & $\beta=\mathbf{1.0}$ & $\beta=5.0$ \\
\cmidrule(l){2-6}
& \multirow{3}{*}{ShuffleMixer}
  & PSNR$\uparrow$ & 34.3912 & \textbf{34.4669} & 34.3857 \\
& & MSE$\downarrow$ & 39.7281 & \textbf{38.7614} & 39.8220 \\
& & SSIM$\uparrow$ & 0.9228  & \textbf{0.9242}  & 0.9228  \\
\cmidrule(l){2-6}
& \multirow{3}{*}{HAT}
  & PSNR$\uparrow$ & 35.0252 & \textbf{35.0089} & 34.7374 \\
& & MSE$\downarrow$ & 32.2430 & \textbf{32.7994} & 34.4470 \\
& & SSIM$\uparrow$ & 0.9331  & \textbf{0.9328}  & 0.9308  \\
\cmidrule(l){2-6}
& \multirow{3}{*}{HiT}
  & PSNR$\uparrow$ & 35.6000 & \textbf{35.7043} & 35.5034 \\
& & MSE$\downarrow$ & 28.9826 & \textbf{28.2897} & 28.7450 \\
& & SSIM$\uparrow$ & 0.9388  & \textbf{0.9403}  & 0.9389  \\
\bottomrule
\end{tabular}}
\label{tab:ablation_kappa_beta}
\end{table}

\section{Conclusion}
\label{sec:conclusion}
In this work, we presented OGG-FR, a plug-and-play optimization framework for UAV infrared image super-resolution. The central challenge is to train lightweight SR models, which are preferred on resource-constrained UAV platforms, using both pixel-domain and frequency-domain supervision even though infrared high-frequency signals often mix weak thermal structures with noise-like residuals. OGG-FR addresses this challenge by decomposing the frequency gradient relative to the pixel gradient and using two update regimes: MGDA-based safe control with variance rectification under destructive conflict and residual-aware orthogonal calibration under compatible optimization. Experiments on the Low-light UAV thermal benchmark demonstrated broad improvements across representative SR models and degradation settings. These findings motivate extending conflict-aware frequency rectification to richer loss combinations and online adaptation on real UAV edge devices.

\bibliography{aaai25}

@article{niqe,
  title={Making a “completely blind” image quality analyzer},
  author={Mittal, Anish and Soundararajan, Rajiv and Bovik, Alan C},
  journal={IEEE Signal processing letters},
  volume={20},
  number={3},
  pages={209--212},
  year={2012},
  publisher={IEEE}
}

@article{kingma2014adam,
  title={Adam: A method for stochastic optimization},
  author={Kingma, Diederik P and Ba, Jimmy},
  journal={arXiv preprint arXiv:1412.6980},
  year={2014}
}

@inproceedings{liang2021swinir,
  title={Swinir: Image restoration using swin transformer},
  author={Liang, Jingyun and others},
  booktitle={Proceedings of the IEEE/CVF ICCV},
  pages={1833--1844},
  year={2021}
}

@article{sun2022shufflemixer,
  title={Shufflemixer: An efficient convnet for image super-resolution},
  author={Sun, Long and others},
  journal={Advances in Neural Information Processing Systems},
  volume={35},
  pages={17314--17326},
  year={2022}
}

@inproceedings{chen2024recursive,
  title={Recursive Generalization Transformer for Image Super-Resolution},
  author={Chen, Zheng and Zhang, Yulun and Gu, Jinjin and Kong, Linghe and Yang, Xiaokang},
  booktitle={ICLR},
  year={2024}
}

@ARTICLE{Chen_2023_CVPR,
  author={Chen, Xiangyu and Wang, Xintao and Zhang, Wenlong and Kong, Xiangtao and Qiao, Yu and Zhou, Jiantao and Dong, Chao},
  journal={IEEE Transactions on Pattern Analysis and Machine Intelligence}, 
  title={HAT: Hybrid Attention Transformer for Image Restoration}, 
  year={2026},
  volume={48},
  number={3},
  pages={2676-2694}}

@inproceedings{agustsson2017ntire,
  title={Ntire 2017 challenge on single image super-resolution: Dataset and study},
  author={Agustsson, Eirikur and Timofte, Radu},
  booktitle={Proceedings of the IEEE conference on computer vision and pattern recognition workshops},
  pages={126--135},
  year={2017}
}

@inproceedings{liu2025catanet,
  title={CATANet: Efficient Content-Aware Token Aggregation for Lightweight Image Super-Resolution},
  author={Liu, Xin and Liu, Jie and Tang, Jie and Wu, Gangshan},
  booktitle={Proceedings of the Computer Vision and Pattern Recognition Conference},
  pages={17902--17912},
  year={2025}
}

@ARTICLE{10938642,
  author={Sun, Chen and others},
  journal={IEEE Transactions on Geoscience and Remote Sensing}, 
  title={NOT-156: Night Object Tracking Using Low-Light and Thermal Infrared: From Multimodal Common-Aperture Camera to Benchmark Datasets}, 
  year={2025},
  volume={63},
  number={},
  pages={1-11}}

@article{zhao2025guidance,
  title={Guidance disentanglement network for Optics-Guided thermal UAV image super-resolution},
  author={Zhao, Zhicheng and others},
  journal={ISPRS Journal of Photogrammetry and Remote Sensing},
  volume={228},
  pages={64--82},
  year={2025},
  publisher={Elsevier}
}

@ARTICLE{11181143,
  author={Huang, Yongsong and others},
  journal={IEEE Journal of Selected Topics in Applied Earth Observations and Remote Sensing}, 
  title={Infrared Image Super-Resolution: A Systematic Review and Future Trends}, 
  year={2025},
  volume={18},
  number={},
  pages={25439-25463}}

@article{wei2024mmpareto,
  title={Mmpareto: Boosting multimodal learning with innocent unimodal assistance},
  author={Wei, Yake and Hu, Di},
  journal={ICML},
  year={2024}
}

@article{sener2018multi,
  title={Multi-task learning as multi-objective optimization},
  author={Sener, Ozan and Koltun, Vladlen},
  journal={Advances in neural information processing systems},
  volume={31},
  year={2018}
}

@article{yu2020gradient,
  title={Gradient surgery for multi-task learning},
  author={Yu, Tianhe and Kumar, Saurabh and Gupta, Abhishek and Levine, Sergey and Hausman, Karol and Finn, Chelsea},
  journal={arXiv preprint arXiv:2001.06782},
  year={2020}
}

@inproceedings{chen2018gradnorm,
  title={Gradnorm: Gradient normalization for adaptive loss balancing in deep multitask networks},
  author={Chen, Zhao and Badrinarayanan, Vijay and Lee, Chen-Yu and Rabinovich, Andrew},
  booktitle={International conference on machine learning},
  pages={794--803},
  year={2018},
  organization={PMLR}
}

@article{liu2023famo,
  title={Famo: Fast adaptive multitask optimization},
  author={Liu, Bo and Feng, Yihao and Stone, Peter and Liu, Qiang},
  journal={Advances in Neural Information Processing Systems},
  volume={36},
  pages={57226--57243},
  year={2023}
}

@article{liu2021conflict,
  title={Conflict-averse gradient descent for multi-task learning},
  author={Liu, Bo and Liu, Xingchao and Jin, Xiaojie and Stone, Peter and Liu, Qiang},
  journal={Advances in neural information processing systems},
  volume={34},
  pages={18878--18890},
  year={2021}
}

@inproceedings{luo2024smfanet,
  title={SMFANet: A lightweight self-modulation feature aggregation network for efficient image super-resolution},
  author={Zheng, Mingjun and Sun, Long and Dong, Jiangxin and Pan, Jinshan},
  booktitle={European conference on computer vision},
  pages={359--375},
  year={2024},
  organization={Springer}
}

@inproceedings{chen2023hit,
  title={HiT-SR: Hierarchical transformer for efficient image super-resolution},
  author={Zhang, Xiang and Zhang, Yulun and Yu, Fisher},
  booktitle={European conference on computer vision},
  pages={483--500},
  year={2024},
  organization={Springer}
}

@inproceedings{li2023craft,
  title={Feature modulation transformer: Cross-refinement of global representation via high-frequency prior for image super-resolution},
  author={Li, Ao and Zhang, Le and Liu, Yun and Zhu, Ce},
  booktitle={Proceedings of the IEEE/CVF international conference on computer vision},
  pages={12514--12524},
  year={2023}
}

@inproceedings{chen2025fusion,
  title={Fusion Meets Diverse Conditions: A High-diversity Benchmark and Baseline for UAV-based Multimodal Object Detection with Condition Cues},
  author={Chen, Chen and Bin, Kangcheng and Hu, Ting and Qi, Jiahao and Liu, Xingyue and Liu, Tianpeng and Liu, Zhen and Liu, Yongxiang and Zhong, Ping},
  booktitle={Proceedings of the IEEE/CVF International Conference on Computer Vision},
  pages={27958--27967},
  year={2025}
}

@article{barbato2026flyawarev2,
  title={FlyAwareV2: A multimodal cross-domain UAV dataset for urban scene understanding},
  author={Barbato, Francesco and Caligiuri, Matteo and Zanuttigh, Pietro},
  journal={Signal Processing: Image Communication},
  pages={117483},
  year={2026},
  publisher={Elsevier}
}

\end{document}